\documentclass{article}
\usepackage{iclr2027_conference,times}
\usepackage[T1]{fontenc}

\usepackage{amsmath}
\usepackage{amssymb}
\usepackage{mathtools}
\usepackage{graphicx}
\usepackage{booktabs}
\usepackage{xcolor}
\usepackage{colortbl}
\usepackage{caption}
\usepackage{algorithm}
\usepackage{algpseudocode}
\usepackage{microtype}
\definecolor{accent}{HTML}{6250D6}
\definecolor{linkblue}{rgb}{0.21,0.49,0.74}
\usepackage[breaklinks,colorlinks,allcolors=linkblue]{hyperref}
\usepackage[capitalize]{cleveref}
\crefname{section}{Sec.}{Secs.}
\Crefname{section}{Section}{Sections}
\Crefname{table}{Table}{Tables}
\crefname{table}{Tab.}{Tabs.}

\algnewcommand{\Input}{\item[\textbf{Input:}]}
\makeatletter
\newcommand{\notice}[1]{\begin{NoHyper}\long\def\@makefntext##1{\noindent\hbox to1.8em{\hss}##1}\let\thefootnote\relax\footnotetext{#1}\end{NoHyper}}
\makeatother

\title{Conditioning Source and Velocity: Feature Inversion for Humans and Machines}

\author{Ashan Perera\thanks{Corresponding author: \texttt{aperera2016@fau.edu}}\hphantom{$^{\dagger}$}, Juan Merlos, Velibor Adzic, Hari Kalva \\
Florida Atlantic University \\
\texttt{\{aperera2016, jmerlosjr2017, vadzic, hkalva\}@fau.edu}}

\iclrfinalcopy
\begin{document}

\setcounter{footnote}{1}
\maketitle
\lhead{}
\notice{This work has been submitted to the IEEE for possible publication. Copyright may be transferred without notice, after which this version may no longer be accessible.}

\begin{center}
\includegraphics[width=\linewidth, trim=0 11bp 0 8bp, clip]{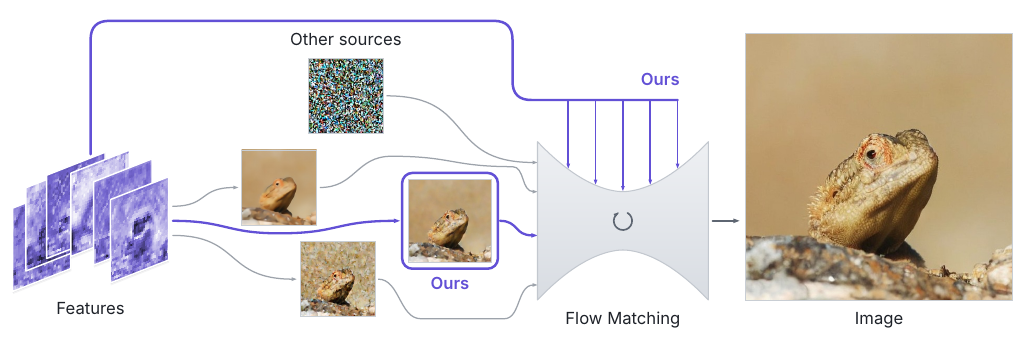}
\captionsetup{hypcap=false}
\captionof{figure}{How well can we recover images from the features of vision models?
To invert features, we predict the source of flow matching from the features and condition its velocity network on them.
Other sources, based on previous works~\citep{lipman2023flow,ren2026fiaflow,ohayon2025pmrf}, are shown for comparison.}
\label{fig:teaser}
\end{center}

\begin{abstract}
In split inference, where a device sends features instead of images, feature inversion is studied as an attack.
We study it as a capability: recovering images that humans can view and off-the-shelf vision models can analyze.
Features are underdetermined representations of input images.
Restoring images from features alone requires generation of plausible details.
Where should generation start, and what should steer it?
Recent black-box methods start from noise unrelated to the features, or start from a predicted point and then generate without features.
With flow matching, we let the features answer both aspects: a feature uncertainty network (FUN) predicts a Gaussian source with a structured covariance, and feature informed transport (FIT) carries its samples to images with a velocity conditioned on the features.
Trained without a perceptual loss, and tested on five networks including ConvNeXt and DINOv3, our method outperforms all baselines in LPIPS, DISTS, and FID, and maintains the best average performance on downstream utility.
The outperformance holds, and in most cases grows, for defended and compressed features.
These results challenge the conventional understanding of the utility of features and their role in privacy preservation.
\end{abstract}

\section{Introduction}
\label{sec:intro}

Split inference runs the first part of a network on a device and sends its features to a server, which runs the rest~\citep{kang2017neurosurgeon,teerapittayanon2017distributed,choi2018deep,choi2018near,eshratifar2019bottlenet,matsubara2022split}.
It is an attractive solution for devices with limited compute, such as IoT cameras, phones and drones.
Consider a camera that sends its features to a server that runs detection.
Split inference also changes what the device shares, since only the features are sent and the server holds no image.
The rest of the network needs no more, so the image does not leave the device and is often taken to be protected~\citep{teerapittayanon2017distributed,matsubara2022split,dong2022privacy,ren2026fiaflow}.

In split inference, feature inversion is studied as an attack on that protection~\citep{he2019model,dong2022privacy,yang2022measuring,yin2023ginver,li2023gan,chen2024dia}.
Recent works show that much of the image can be recovered from the features~\citep{lei2025drag,zhang2025unlocking,leggio2026position,ren2026fiaflow}.
However, feature inversion has not always been studied as an attack.
Some of the earliest works used it for interpretability, to understand how much of the image information remains in the features~\citep{mahendran2015understanding,dosovitskiy2016inverting}.
Later works have continued this line of research, but the main focus has shifted away from it~\citep{nash2019inverting,rombach2020making,neukirch2025featinv,rathjens2026understanding}.

Beyond interpretability, we recognize two beneficial uses of inversion in split inference: viewing by humans and analysis by different vision models.
The first applies whenever a viewer wants to see the image and only the features are available, for example to check an alert from the detector~\citep{choi2018near}.
The second runs an off-the-shelf vision model on the recovered image for new tasks, so the features in effect serve new models that they were not designed for.
Inversion is therefore a capability, and who uses it decides whether it is an attack or a service.

White-box methods need the network's architecture and weights, and they often optimize each image separately~\citep{mahendran2015understanding,he2019model,dong2022privacy,li2023gan,lei2025drag,zhang2025unlocking}.
A black-box inversion model needs only pairs of images and features, and once trained it inverts new features without further optimization.
We consider feature inversion in the black-box setting (\cref{sec:bg-inversion}), which is well studied~\citep{dosovitskiy2016inverting,he2019model,yang2022measuring,chen2024dia,zhang2025unlocking,neukirch2025featinv,ren2026fiaflow} and assumes little about the network on the device.
The same approach applies to networks of different kinds, including those whose internals are not disclosed.

Network operations are usually lossy, so the original image cannot be recovered exactly from its features.
In the absence of exact reconstruction, what should we aim for?
We believe the most useful image is one that maximizes utility for humans and vision models.

Such an image contains detail that the features do not determine, so this detail has to be generated.
In the black-box setting, recent generative methods start from noise that does not depend on the features~\citep{chen2024dia,neukirch2025featinv} or from a single point predicted from the features~\citep{zhang2025unlocking,ren2026fiaflow}.
Neither starting point accounts for what the features leave uncertain, so we instead predict a Gaussian that estimates the center and the uncertainty of the images with these features.
We then use flow matching to carry a draw from this Gaussian to one of these images, with a velocity network that is also conditioned on the features (\cref{fig:teaser}).

\section{Related Work}
\label{sec:related}

\paragraph{Feature inversion.}
An inversion model trained on pairs of images and features with a squared error on pixels~\citep{dosovitskiy2016inverting,he2019model} returns approximately the mean of the images that have the same features, which is blurry wherever these images differ~\citep{dosovitskiy2016inverting}.
Generative inversion models can return several samples for the same features~\citep{nash2019inverting,rombach2020making,shocher2020semantic}.
Among those trained only on pairs, DIA~\citep{chen2024dia} trains a diffusion model conditioned on the features, and FeatInv~\citep{neukirch2025featinv} adds a ControlNet~\citep{zhang2023controlnet} to a pretrained diffusion model.
Both start from standard Gaussian noise.
In contrast, DMB~\citep[in its black-box variant]{zhang2025unlocking} and FIA-Flow~\citep{ren2026fiaflow} predict the initial latent of a pretrained model from the features and do not condition the model on them.
Closest to our work, FIA-Flow maps this latent to an image with flow matching in one step and without noise, so it returns a single image for given features.
We condition our model on the features and also use them to predict its source, the distribution it starts from.

\paragraph{Condition-dependent sources.}
Sources that depend on the condition are common in image restoration and editing~\citep{meng2022sdedit,liu2023i2sb,yue2023resshift}.
PMRF~\citep{ohayon2025pmrf} regresses the mean of the clean images for an input and starts a rectified flow from this mean plus noise of one fixed level.
UPSR~\citep{zhang2025uncertainty} initializes super-resolution from the input plus noise and weights this noise at each pixel according to an estimated uncertainty.
The variance of the source tends to collapse when it is learned jointly with the velocity network, so CAR-Flow~\citep{chen2025carflow} learns only a shift of the source, while CrossFlow~\citep{liu2025flowing} and CSFM~\citep{kim2026better} regularize the variance toward one.
\citet{issachar2025designing} instead take the mean and covariance of a Gaussian source from the data of each class, and for a text prompt they regress the mean and fix the covariance.
We predict the covariance from the features and fit it using maximum likelihood with the mean fixed, following \citet{dorta2018structured}.

\section{Background}
\label{sec:background}

\subsection{Black-Box Feature Inversion}
\label{sec:bg-inversion}

A fixed network $h$ maps an image $x \in \mathbb{R}^{H \times W \times C}$ to features $f = h(x) \in \mathbb{R}^{H_f \times W_f \times C_f}$.
Both are spatial maps, grids of positions with several values at each position.
We consider networks whose features share the layout of the image.
That is, the feature at a position of $f$ describes the part of the image at the same relative location.

We study recovering $x$ from $f$ when $h$ is available only through queries, which evaluate $h$ on chosen images without exposing its parameters or gradients.
The queries are made once, before training.
Evaluating $h$ on $N$ images gives the paired set $S = \{(x_i, f_i)\}_{i=1}^{N}$ with $f_i = h(x_i)$, after which $h$ is not evaluated again.

An inversion model $g$ is trained on $S$.
Given the features $f = h(x)$ of an image $x$ outside $S$, it returns an estimate $\hat{x} = g(f)$.
In the black-box setting, $g$ knows neither the architecture of $h$ nor the layer that produces $f$.

\subsection{Flow Matching}
\label{sec:bg-flow}

Flow matching~\citep{lipman2023flow,liu2023rectified,albergo2023interpolants} trains a velocity field that carries samples of a source distribution $p_0$ to samples of a target distribution $p_1$.
Training pairs $(z_0, z_1)$ are drawn from a coupling, which may be any joint distribution with marginals $p_0$ and $p_1$~\citep{liu2023rectified,tong2024improving}.
We focus on rectified flow~\citep{liu2023rectified}, in which each pair is joined by the linear path
\begin{equation}
  z_t = (1 - t)\, z_0 + t\, z_1, \qquad t \in [0, 1],
  \label{eq:path}
\end{equation}
so $z_t$ moves from $z_0$ to $z_1$ with constant velocity $z_1 - z_0$.
A network $v_\theta(z_t, t)$ learns this velocity by minimizing
\begin{equation}
  \mathcal{L}_{\mathrm{FM}}(\theta) = \mathbb{E}_{(z_0, z_1),\, t} \big\lVert v_\theta(z_t, t) - (z_1 - z_0) \big\rVert^2
  \label{eq:fm}
\end{equation}
with $t \sim \mathcal{U}[0, 1]$.

The target $z_1 - z_0$ is not determined by $(z_t, t)$, because the paths of many pairs typically pass through the same point $z_t$.
The minimizer of \cref{eq:fm} is therefore the average over these pairs,
\begin{equation}
  v^\ast(z, t) = \mathbb{E}\big[z_1 - z_0 \mid z_t = z\big].
  \label{eq:vstar}
\end{equation}
That is, $v^\ast$ is the mean direction of the paths through $z$ at time $t$.
Assuming $v^\ast$ is locally bounded and the ODE
\begin{equation}
  \frac{\mathrm{d} z_t}{\mathrm{d} t} = v^\ast(z_t, t), \qquad z_0 \sim p_0,
  \label{eq:ode}
\end{equation}
has a unique solution, integrating it from $t = 0$ to $t = 1$ gives $z_1 \sim p_1$ for any choice of $p_0$ and coupling~\citep{liu2023rectified}.

\subsection{Latent Space}
\label{sec:bg-latent}

We work in the latent space of a fixed autoencoder.
Its encoder $\mathcal{E}$ maps an image $x$ to the latent $z_1 = \mathcal{E}(x)$, and its decoder $\mathcal{D}$ maps latents back to images.
The target distribution $p_1$ is the distribution of the latents of images.
The latent is a spatial map as well, $z_1 \in \mathbb{R}^{H_z \times W_z \times C_z}$, with $d = H_z W_z C_z$ values.
Like the features, it shares the layout of the image.
Although the three maps differ in resolution, the local information remains in the positional neighborhood of a given location.

\section{Method}
\label{sec:method}

We invert features $f$ in two stages (\cref{fig:teaser}).
The first is the feature uncertainty network (FUN, \cref{sec:fun}), which predicts a Gaussian that gives the center and the uncertainty of the images with features $f$.
The second is the feature informed transport (FIT, \cref{sec:fit}), which carries a draw from this Gaussian to one of these images.

\subsection{Feature Uncertainty Network}
\label{sec:fun}

The network $h$ is trained for its own task, which does not require its features to preserve the image.
In general, distinct local details in an image are mapped into similar values in the feature space.
These images occupy a region of image space, and the extent of this region is the ``uncertainty'' that $f$ leaves about the image.
The uncertainty is zero if $h$ keeps everything, and if $h$ keeps nothing the region contains every image.
What $h$ keeps depends on the image and on the position, so what $f$ leaves uncertain depends on both as well.

A detail that the features leave ambiguous within a part of the image, such as the exact course of an edge, spans several nearby positions, so across the images with features $f$ the image values at these positions vary together.
A separate uncertainty for each position treats these values as independent and overstates the size of the region, since values that vary together fill less of image space than independent values that vary by the same amount.
An accurate measure of the uncertainty therefore includes the covariance between nearby values~\citep{nehme2023uncertainty}.

FUN treats the prediction of the image from $f$ as a heteroscedastic regression problem~\citep{kendall2017what}.
It predicts a Gaussian whose mean and covariance are the learned center and uncertainty.
The Gaussian summarizes the images with features $f$ by these two statistics, while we do not assume that the distribution of these images is Gaussian.
We break FUN into two steps, one to learn the mean and another to learn the covariance.

\paragraph{Conditional mean.}
The first step of FUN predicts the latent $z_1$ of an image from its features $f$ with a network $\mu_\phi$, which we train by minimizing
\begin{equation}
  \mathcal{L}_{\mu}(\phi) = \mathbb{E}_{(x, f)} \big\lVert \mu_\phi(f) - z_1 \big\rVert^2,
  \label{eq:mean-loss}
\end{equation}
where the expectation over images and their features is estimated on the pairs in $S$.
By the same argument as for \cref{eq:vstar}, the minimizer of \cref{eq:mean-loss} is the conditional mean
\begin{equation}
  \mu^\ast(f) = \mathbb{E}\big[z_1 \mid f\big].
  \label{eq:mustar}
\end{equation}
That is, $\mu^\ast(f)$ is the average latent of the images whose features are $f$.
Any suitable estimator $\mu_\phi$ can be used in the later step.

\paragraph{Conditional covariance.}
The next step of FUN predicts how the latents of the images with features $f$ spread around $\mu_\phi(f)$.
We treat latents as vectors in $\mathbb{R}^{d}$ with their values listed position by position in raster order, and model $z_1$ given $f$ as the Gaussian $\mathcal{N}\big(\mu_\phi(f), \Sigma_\psi(f)\big)$.
Its covariance $\Sigma = \Sigma_\psi(f)$ is predicted by a network with parameters $\psi$.
Following \citet{dorta2018structured}, we fix $\phi$ and train $\psi$ by minimizing the negative log-likelihood
\begin{equation}
  \mathcal{L}_{\Sigma}(\psi) = \mathbb{E}_{(x, f)} \Big[ \tfrac{1}{2}\, r^\top \Sigma^{-1} r + \tfrac{1}{2} \log\det\Sigma \Big],
  \label{eq:cov-nll}
\end{equation}
where $r = z_1 - \mu_\phi(f)$ is the residual of the mean and constants are dropped.
For each $f$, the minimizer of \cref{eq:cov-nll} over all symmetric positive-definite $\Sigma$ is $\mathbb{E}\big[r r^\top \mid f\big]$.
That is, the optimal $\Sigma$ is the conditional covariance of $z_1$ given $f$ when $\mu_\phi = \mu^\ast$, and any error of $\mu_\phi$ adds to it.

\begin{figure}[t]
\centering
\includegraphics{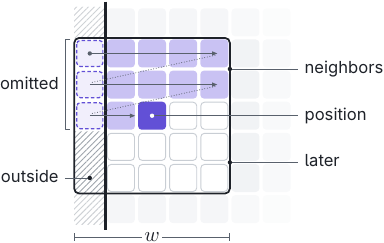}
\caption{Neighbors of a position near the left border of the latent, for a window of size $w = 5$.
Of the 12 window positions that precede it in raster order, 3 fall outside the latent and are omitted.}
\label{fig:neighbors}
\end{figure}

Predicting $\Sigma$ entry by entry is impractical.
Its free entries, whose number grows quadratically with $d$, must form a positive-definite matrix, and \cref{eq:cov-nll} requires the inverse and the determinant of that matrix.
A diagonal $\Sigma$ avoids these problems, but it makes the values of a sample independent.

We therefore predict $\Sigma$ through a factor,
\begin{equation}
  \Sigma = A A^\top,
  \label{eq:cov-factor}
\end{equation}
where $A \in \mathbb{R}^{d \times d}$ is lower triangular with a positive diagonal.
Every such $A$ gives a positive-definite $\Sigma$ and is its Cholesky factor.
Substituting \cref{eq:cov-factor} into \cref{eq:cov-nll} gives the loss we minimize,
\begin{equation}
  \mathcal{L}_{\Sigma}(\psi) = \mathbb{E}_{(x, f)} \Big[ \tfrac{1}{2} \big\lVert A^{-1} r \big\rVert^2 + \sum_{k=1}^{d} \log A_{kk} \Big],
  \label{eq:cov-loss}
\end{equation}
since $\log\det\Sigma = 2 \sum_{k} \log A_{kk}$ and $r^\top \Sigma^{-1} r = \lVert A^{-1} r \rVert^2$.
Neither $\Sigma$ nor its inverse is formed, and $A^{-1} r$ takes one forward substitution.
A dense $A$, however, still has free entries whose number grows quadratically with $d$.

We restrict the model to correlations between nearby values by making $A$ sparse.
The neighbors of a position of the latent are the positions of the $w \times w$ window around it that precede it in raster order, as a lower-triangular $A$ requires.
\Cref{fig:neighbors} shows the neighbors of a position near the border, where the window positions outside the latent are omitted.
An entry $A_{kj}$ can be nonzero only if the $j$-th latent value lies at the position of the $k$-th or at one of its neighbors.
The sparse $A$ has free entries whose number grows linearly with $d$, and only these are predicted.
\citet{dorta2018structured} use such a pattern for the Cholesky factor of the precision matrix, whereas we factor the covariance, so that drawing from the Gaussian takes one product with $A$.
Over the resulting family of covariances, minimizing \cref{eq:cov-nll} is equivalent, for each $f$, to minimizing $\mathrm{KL}\big(\mathcal{N}(0, \mathbb{E}[r r^\top \mid f]) \,\Vert\, \mathcal{N}(0, \Sigma)\big)$.
The diagonal covariance is the member of this family in which every entry off the diagonal of $A$ is zero.

\paragraph{Output.}
For the diagonal covariance, the standard deviation of the $k$-th latent value is $A_{kk}$, and in general it is the norm of the $k$-th row of $A$,
\begin{equation}
  \sigma_k = \sqrt{\sum_{j} A_{kj}^2}.
  \label{eq:cov-std}
\end{equation}
Arranged as the latent, these marginal standard deviations form the uncertainty map $\sigma$ of $f$.
For features $f$, the output of FUN is the Gaussian $\mathcal{N}\big(\mu_\phi(f), \Sigma_\psi(f)\big)$ and its uncertainty map $\sigma$.

\subsection{Feature Informed Transport}
\label{sec:fit}

FUN now provides the source distribution for flow matching.
Let $p_1(\cdot \mid f)$ denote the distribution of the latents of the images with features $f$.
FIT uses flow matching to carry a draw from the Gaussian onto these latents.
The features inform the transport twice, through its source, which FUN predicts from $f$, and through its velocity, which we condition on $f$.

\paragraph{Source.}
Flow matching proceeds as in \cref{sec:bg-flow}, with a source that depends on $f$.
For each $f$, the source is the Gaussian of FUN, and a draw from it is
\begin{equation}
  z_0 = \underbracket[0.4pt][2pt]{\mu_\phi(f)}_{\text{center}} + \underbracket[0.4pt][2pt]{A\,\varepsilon\vphantom{\mu_\phi(f)}}_{\text{spread}}, \qquad \varepsilon \sim \mathcal{N}(0, I).
  \label{eq:fit-source}
\end{equation}
With $\mu_\phi = 0$ and $A = I$ the source is $\mathcal{N}(0, I)$, and with $A = 0$ it is the mean alone.
A factor $A = sI$ with one scalar $s$ for all $f$ gives the mean plus noise of one level~\citep{ohayon2025pmrf,issachar2025designing}.
The source $\mathcal{N}(0, I)$ does not depend on $f$, whereas the Gaussian of FUN already has the predicted center and spread of $p_1(\cdot \mid f)$.
The mean alone has no spread and gives one image for $f$, but many images have these features, so the output has to be a sample.
One noise level gives every $f$ and every value the same spread, although what $f$ leaves uncertain depends on the image and on the position.
For training, $z_1$ is the latent of an image and $z_0$ is a draw from FUN for the features of the same image, with $\varepsilon$ drawn independently of the image.

\begin{algorithm}[t]
\caption{Sampling an image from features}
\label{alg:sampling}
\begin{algorithmic}[1]
\Require FUN networks $\mu_\phi$ and $\Sigma_\psi$, velocity network $v_\theta$, decoder $\mathcal{D}$
\Input features $f$
\State $\mu \gets \mu_\phi(f)$
\State $A \gets$ factor of $\Sigma_\psi(f)$, with $\Sigma_\psi(f) = A A^\top$
\State $\sigma \gets$ uncertainty map, with $\sigma_k$ the norm of row $k$ of $A$
\State $\varepsilon \sim \mathcal{N}(0, I)$
\State $z_0 \gets \mu + A\,\varepsilon$
\State $\hat{z}_1 \gets$ solve $\frac{\mathrm{d}z_t}{\mathrm{d}t} = v_\theta(z_t, t; f, \sigma)$ from $t = 0$ to $t = 1$
\State \Return $\hat{x} = \mathcal{D}(\hat{z}_1)$
\end{algorithmic}
\end{algorithm}

\paragraph{Conditional velocity.}
In PMRF~\citep{ohayon2025pmrf} and FIA-Flow~\citep{ren2026fiaflow}, the source depends on the input and the velocity does not.
With a velocity $v_\theta(z_t, t)$, the result of \cref{sec:bg-flow} covers all $f$ together: the ODE of \cref{eq:ode} carries the draws for all $f$, pooled into one source $p_0$, onto $p_1$.
The result does not imply that a draw for $f$ ends on $p_1(\cdot \mid f)$.
We therefore give $f$ to the velocity, $v_\theta(z_t, t; f)$, so that the result applies to each $f$ separately.

With $\mathcal{N}(0, I)$ as the source, the noise part of $z_t$ is $(1 - t)\,\varepsilon$ by \cref{eq:path}, with standard deviation $1 - t$ at every value, a function of $t$ alone.
\Cref{eq:fit-source} instead makes the noise part $(1 - t) A\varepsilon$, with standard deviation $(1 - t)\,\sigma_k$ at the $k$-th value.
The standard deviation now also depends on $\sigma_k$, so we condition the velocity on the uncertainty map, which informs the velocity network of the starting noise.
In principle $f$ determines $\sigma$, but FUN has already computed it, so we pass it and write the velocity as $v_\theta(z_t, t; f, \sigma)$.

The loss of FIT is
\begin{equation}
  \mathcal{L}_{\mathrm{FIT}}(\theta) = \mathbb{E}_{(x, f),\, \varepsilon,\, t} \big\lVert v_\theta(z_t, t; f, \sigma) - (z_1 - z_0) \big\rVert^2,
  \label{eq:fit-loss}
\end{equation}
with $z_0$ from \cref{eq:fit-source}, $z_t$ from \cref{eq:path} and $t \sim \mathcal{U}[0, 1]$.
FUN stays fixed while FIT is trained, so $\sigma$ remains a function of $f$ and passing it leaves the minimizer unchanged.
As with \cref{eq:vstar}, the minimizer of \cref{eq:fit-loss} is
\begin{equation}
  v^\ast(z, t; f) = \mathbb{E}\big[z_1 - z_0 \mid z_t = z,\, f\big].
  \label{eq:fit-vstar}
\end{equation}
Here the mean direction of the paths is taken only over the images with features $f$.
By the result of \cref{sec:bg-flow}, applied for each $f$ under the same assumptions, the ODE with $v^\ast(z_t, t; f)$ carries the Gaussian for $f$ onto $p_1(\cdot \mid f)$.
The derivation so far does not depend on how the velocity network is built.

\paragraph{Velocity network.}
The transport has to supply the image detail that $f$ leaves open, and generative models pretrained on many images have learned to generate it.
FIT is designed to support pretrained models that operate in the latent space of \cref{sec:bg-latent}.
If we initialize $v_\theta$ with one, it has to be adapted, since such a model was not trained for our source and has no input for $f$.

The usual condition of such models is a text prompt~\citep{rombach2022ldm,esser2024scaling,chen2024pixart,xie2025sana}, which is not a spatial map.
Every position of an activation map receives the same prompt, so conditioning on $f$ in this way would ignore where a feature lies.
The features share the layout of the image and need a scheme that acts position by position.
As StableSR~\citep{wang2024exploiting} and SUPIR~\citep{yu2024scaling} do for restoration, we insert layers similar to SFT~\citep{wang2018sft}, each of which turns an activation map $u$ of the velocity network into
\begin{equation}
  u' = u + \gamma \odot u + \beta,
  \label{eq:fit-layer}
\end{equation}
where $\gamma$ and $\beta$ are maps predicted from $f$ by a small network, at the resolution of $u$, and $\odot$ is the elementwise product.
SFT computes $\gamma \odot u + \beta$, and with the added $u$ the layer leaves $u$ unchanged when $\gamma$ and $\beta$ are zero.
We apply \cref{eq:fit-layer} at various resolutions of $v_\theta$.

We train the new layers jointly with the pretrained model.
Randomly initialized, they would add noise to its activations from the first training step, which can damage what the model has learned~\citep{zhang2023controlnet}.
To avoid this, we initialize the layers that output $\gamma$ and $\beta$ to zero, as in ControlNet~\citep{zhang2023controlnet}.
The uncertainty map has the shape of the latent, so it is passed with $z_t$ to the first layer of $v_\theta$, through a new layer that is also initialized to zero.
As a result, $v_\theta$ starts as the pretrained model.

\paragraph{Sampling.}
\Cref{alg:sampling} concludes the method by sampling an image from features.
This procedure is the inversion model $g$ of \cref{sec:bg-inversion}, and its estimate $\hat{x}$ is a sample, since each draw of $\varepsilon$ gives another image for the same $f$.

\section{Experiments and Results}
\label{sec:results}

\begin{table}[t]
  \centering
  \caption{Feature inversion results on five networks. The best values are in bold.}
  \label{tab:main-results}
  \footnotesize
  \setlength{\tabcolsep}{1.5pt}
  \resizebox{\linewidth}{!}{%
  \begin{tabular}{lr>{\hspace{1pt}}ccccc>{\hspace{1pt}}ccccc>{\hspace{1pt}}ccccc}
    \toprule
    & & \multicolumn{5}{c}{\textbf{PSNR $\uparrow$}} & \multicolumn{5}{c}{\textbf{LPIPS $\downarrow$}} & \multicolumn{5}{c}{\textbf{DISTS $\downarrow$}} \\
    \cmidrule(lr){3-7} \cmidrule(lr){8-12} \cmidrule(lr){13-17}
    \textbf{Method} & \textbf{NFE} & ConvNeXt & Swin & DINOv3 & SigLIP~2 & PE & ConvNeXt & Swin & DINOv3 & SigLIP~2 & PE & ConvNeXt & Swin & DINOv3 & SigLIP~2 & PE \\
    \midrule
    DO & 1 & 20.85 & 21.23 & 20.02 & 19.15 & 21.13 & 0.648 & 0.638 & 0.678 & 0.674 & 0.637 & 0.395 & 0.392 & 0.407 & 0.412 & 0.392 \\
    DB & 1 & 20.81 & 21.35 & 20.07 & 19.25 & 21.08 & 0.654 & 0.631 & 0.677 & 0.683 & 0.643 & 0.404 & 0.394 & 0.407 & 0.425 & 0.390 \\
    FIA-Align & 1 & 21.79 & \textbf{22.53} & 21.26 & 20.84 & \textbf{22.78} & 0.431 & 0.411 & 0.461 & 0.431 & 0.390 & 0.317 & 0.304 & 0.333 & 0.317 & 0.292 \\
    FIA-Flow & 1 & \textbf{21.86} & 22.41 & \textbf{21.35} & \textbf{21.01} & 22.75 & 0.201 & 0.181 & 0.218 & 0.213 & 0.176 & 0.125 & 0.114 & 0.137 & 0.132 & 0.114 \\
     & 5 & 21.20 & 21.72 & 20.85 & 20.46 & 22.08 & 0.186 & 0.168 & 0.207 & 0.199 & 0.163 & 0.115 & 0.106 & 0.128 & 0.121 & 0.108 \\
     & 20 & 20.89 & 21.53 & 20.57 & 20.20 & 21.83 & 0.195 & 0.175 & 0.215 & 0.209 & 0.172 & 0.120 & 0.108 & 0.134 & 0.125 & 0.113 \\
    FeatInv & 20 & 15.17 & 16.13 & 15.75 & 14.50 & 16.16 & 0.346 & 0.301 & 0.375 & 0.391 & 0.306 & 0.202 & 0.179 & 0.217 & 0.219 & 0.189 \\
    \rowcolor{accent!12} \hspace{-\tabcolsep}{\color{accent}\vrule width 1.5pt}\hspace{2.5pt}\textbf{Ours} & 20 & 20.78 & 21.27 & 20.57 & 20.20 & 21.81 & \textbf{0.171} & \textbf{0.159} & \textbf{0.193} & \textbf{0.197} & \textbf{0.154} & \textbf{0.098} & \textbf{0.092} & \textbf{0.115} & \textbf{0.112} & \textbf{0.094} \\
    \midrule
    & & \multicolumn{5}{c}{\textbf{FID $\downarrow$}} & \multicolumn{5}{c}{\textbf{MUSIQ $\uparrow$}} & \multicolumn{5}{c}{\textbf{{\boldmath$f$} distance $\downarrow$}} \\
    \cmidrule(lr){3-7} \cmidrule(lr){8-12} \cmidrule(lr){13-17}
    \textbf{Method} & \textbf{NFE} & ConvNeXt & Swin & DINOv3 & SigLIP~2 & PE & ConvNeXt & Swin & DINOv3 & SigLIP~2 & PE & ConvNeXt & Swin & DINOv3 & SigLIP~2 & PE \\
    \midrule
    DO & 1 & 106.17 & 106.12 & 114.92 & 127.04 & 110.26 & 20.75 & 21.30 & 19.34 & 19.30 & 21.86 & 0.208 & 0.763 & 0.278 & 0.569 & 0.439 \\
    DB & 1 & 111.70 & 106.17 & 117.51 & 129.61 & 103.82 & 20.86 & 22.39 & 19.92 & 18.78 & 20.91 & 0.209 & 0.755 & 0.289 & 0.571 & 0.429 \\
    FIA-Align & 1 & \phantom{0}63.58 & \phantom{0}52.49 & \phantom{0}64.68 & \phantom{0}58.90 & \phantom{0}46.90 & 50.44 & 50.67 & 47.01 & 50.79 & 52.50 & 0.122 & 0.667 & 0.148 & 0.424 & 0.225 \\
    FIA-Flow & 1 & \phantom{00}6.44 & \phantom{00}5.80 & \phantom{00}7.08 & \phantom{00}7.11 & \phantom{00}5.80 & 66.51 & 67.07 & 67.26 & 67.32 & 66.96 & 0.054 & 0.567 & 0.051 & \textbf{0.282} & 0.116 \\
     & 5 & \phantom{00}4.90 & \phantom{00}4.25 & \phantom{00}5.40 & \phantom{00}5.26 & \phantom{00}4.40 & 68.02 & 68.19 & 68.23 & 68.72 & 68.30 & 0.055 & \textbf{0.566} & 0.052 & 0.284 & 0.114 \\
     & 20 & \phantom{00}5.14 & \phantom{00}4.35 & \phantom{00}5.85 & \phantom{00}5.58 & \phantom{00}4.71 & 67.84 & 67.89 & 67.66 & 68.54 & 67.97 & 0.061 & 0.576 & 0.062 & 0.298 & 0.122 \\
    FeatInv & 20 & \phantom{00}8.74 & \phantom{00}7.08 & \phantom{00}8.80 & \phantom{00}9.78 & \phantom{00}7.23 & \textbf{71.94} & \textbf{71.48} & \textbf{70.48} & \textbf{72.24} & \textbf{71.39} & 0.073 & 0.654 & 0.063 & 0.350 & 0.171 \\
    \rowcolor{accent!12} \hspace{-\tabcolsep}{\color{accent}\vrule width 1.5pt}\hspace{2.5pt}\textbf{Ours} & 20 & \phantom{00}\textbf{4.19} & \phantom{00}\textbf{3.88} & \phantom{00}\textbf{4.49} & \phantom{00}\textbf{4.98} & \phantom{00}\textbf{3.86} & 68.27 & 68.25 & 68.41 & 68.58 & 67.96 & \textbf{0.052} & 0.591 & \textbf{0.041} & \textbf{0.282} & \textbf{0.113} \\
    \midrule
    & & \multicolumn{5}{c}{\textbf{CLIP agreement (\%) $\uparrow$}} & \multicolumn{5}{c}{\textbf{Segmentation mAP (\%) $\uparrow$}} & \multicolumn{5}{c}{\textbf{LVLM-C (\%) $\uparrow$}} \\
    \cmidrule(lr){3-7} \cmidrule(lr){8-12} \cmidrule(lr){13-17}
    \textbf{Method} & \textbf{NFE} & ConvNeXt & Swin & DINOv3 & SigLIP~2 & PE & ConvNeXt & Swin & DINOv3 & SigLIP~2 & PE & ConvNeXt & Swin & DINOv3 & SigLIP~2 & PE \\
    \midrule
    DO & 1 & 24.1 & 23.6 & 22.2 & 18.9 & 23.2 & 11.7 & 11.1 & \phantom{0}9.5 & \phantom{0}8.4 & \phantom{0}9.7 & 39.0 & 39.4 & 35.8 & 33.9 & 36.6 \\
    DB & 1 & 24.9 & 25.3 & 22.4 & 19.4 & 27.1 & \phantom{0}9.4 & 11.3 & \phantom{0}8.3 & \phantom{0}7.9 & 10.9 & 38.6 & 39.4 & 37.5 & 35.6 & 39.7 \\
    FIA-Align & 1 & 45.1 & 50.6 & 43.4 & 49.1 & 53.1 & 29.4 & 32.4 & 26.0 & 32.0 & 33.1 & 61.0 & 64.3 & 58.7 & 59.9 & 65.0 \\
    FIA-Flow & 1 & 77.5 & 79.4 & 76.1 & 76.9 & 79.0 & 62.7 & 66.8 & 62.1 & 66.1 & 65.3 & 72.4 & 73.8 & 70.9 & 69.9 & 73.7 \\
     & 5 & 79.1 & 80.4 & 77.4 & 78.2 & 80.7 & 62.3 & 66.2 & 60.4 & 65.0 & 66.2 & 72.7 & 73.7 & 70.8 & 70.8 & 73.8 \\
     & 20 & 77.8 & 79.7 & 76.3 & 77.4 & 79.5 & 58.2 & 63.8 & 56.1 & 60.4 & 61.4 & 72.1 & 74.3 & 70.3 & 70.2 & 73.8 \\
    FeatInv & 20 & 72.4 & 75.1 & 72.0 & 70.7 & 75.3 & 52.7 & 55.5 & 51.6 & 50.2 & 53.6 & 61.7 & 66.8 & 58.4 & 58.7 & 65.9 \\
    \rowcolor{accent!12} \hspace{-\tabcolsep}{\color{accent}\vrule width 1.5pt}\hspace{2.5pt}\textbf{Ours} & 20 & \textbf{80.6} & \textbf{81.4} & \textbf{79.7} & \textbf{79.3} & \textbf{81.2} & \textbf{65.9} & \textbf{69.1} & \textbf{64.5} & \textbf{67.2} & \textbf{67.1} & \textbf{74.5} & \textbf{75.3} & \textbf{72.0} & \textbf{73.1} & \textbf{74.8} \\
    \bottomrule
  \end{tabular}}
\end{table}

To cover modern networks of different kinds, we test our method on five networks $h$ whose training spans image classification (ConvNeXt-B, \citealp{liu2022convnext}; Swin-B, \citealp{liu2021swin}), self-supervised learning (DINOv3 ViT-B/16, \citealp{simeoni2026dinov3}) and image-text pretraining (SigLIP~2 L/16, \citealp{tschannen2025siglip2}; PE-Spatial B/16, \citealp{bolya2025perception}).
The images are $512^2$ and come from ImageNet~\citep{russakovsky2015imagenet}, with $N = 68{,}410$ for training and $4{,}474$ for testing.
For FUN we use a window of size $w = 5$ (as depicted in \cref{fig:neighbors}), for FIT we initialize $v_\theta$ with Stable Diffusion~2.1 base~\citep[SD2.1;][]{rombach2022ldm}, and we solve the ODE of \cref{alg:sampling} with Heun's method.
We compare against baselines from recent feature inversion works: DO and DB~\citep{zhang2025unlocking}, FeatInv~\citep{neukirch2025featinv}, and FIA-Align and FIA-Flow~\citep{ren2026fiaflow}.
To isolate the contribution of each method from that of its training setup, we reimplement the baselines and train all methods in a shared setting.
For example, we finetune the SD2.1 of FIA-Flow as we do in FIT, instead of training a LoRA~\citep{hu2022lora} as originally described, since we observe that this improves FIA-Flow on all five networks.
For stochastic methods, such as FeatInv, our method and its ablations, we take five samples per $f$ and report the mean of every metric over the five.
The default number of function evaluations (NFE) is 1 for FIA-Flow (the default in~\citealp{ren2026fiaflow}) and 20 for both FeatInv and our method.

\Cref{tab:main-results} presents the feature inversion results.
We evaluate pixel fidelity (PSNR) and human perception (LPIPS, \citealp{zhang2018lpips}; DISTS, \citealp{ding2022dists}; FID, \citealp{heusel2017fid}; MUSIQ, \citealp{ke2021musiq}).
As a measure of consistency with the features, we apply $h$ to the recovered image $\hat{x}$ and report the cosine distance between $h(\hat{x})$ and $f$ ($f$ distance).
To test the analysis by vision models (\cref{sec:intro}), we also evaluate utility on $\hat{x}$ with models that have never seen $h$ or its features (CLIP agreement, \citealp{radford2021clip}; segmentation mAP, \citealp{lin2014coco}; and LVLM-C with a vision-language model, \citealp{ren2026fiaflow}).

\begin{figure}[t]
\centering
\includegraphics[width=\linewidth]{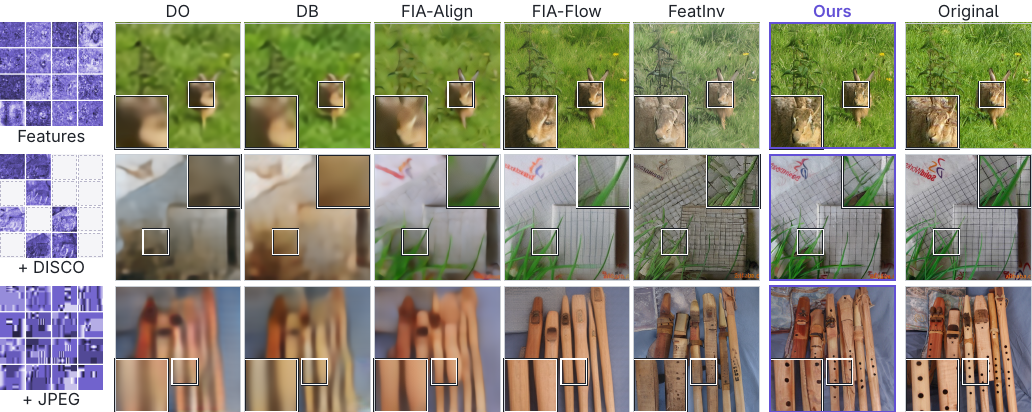}
\caption{Examples of feature inversion on ConvNeXt.
\textit{Top row.} Inversion of the original features.
\textit{Middle row.} Inversion after applying the DISCO defense.
\textit{Bottom row.} Inversion after applying heavy JPEG compression.}
\label{fig:main-examples}
\end{figure}

The highest PSNR on every network belongs to FIA-Flow at NFE 1 or to FIA-Align.
This is expected from the perception-distortion tradeoff~\citep{blau2018perception}, by which the cost of achieving lower distortion has to be paid in terms of perceptual quality.
By the criterion of \cref{sec:intro}, the most useful image is one that humans and vision models can act on as they would the original, and the metrics of human perception and utility are better measures of this.
FeatInv scores highest on MUSIQ on every network.
MUSIQ rates an image without reference to the original, and on LPIPS and DISTS, which do refer to the original, the values of FeatInv are roughly double ours.
On all other metrics of human perception and utility, our method scores best on every network.
Interestingly, although FIA-Flow trains with a loss from the LPIPS family, our method reaches a lower LPIPS on every network without a perceptual loss.
\Cref{fig:main-examples} (top row) compares the recovered images on ConvNeXt.

\begin{figure}[t]
\centering
\includegraphics{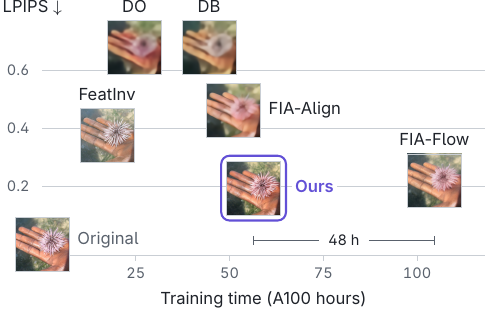}
\caption{Training time against LPIPS on DINOv3.
Each mark is a crop of the image that the method recovers for one test image.}
\label{fig:train-time}
\end{figure}

In our shared setting, we also observe a large reduction in training time against FIA-Flow across all five networks.
\Cref{fig:train-time} shows the case of DINOv3, where our method also has a lower LPIPS than FIA-Flow.
Both methods are based on flow matching, and we attribute our shorter training time to the fact that our method avoids decoding and calculation of any perceptual losses.

\subsection{Ablations}
\label{sec:ablations}

\begin{table}[t]
  \centering
  \caption{Ablations on ConvNeXt. As in \cref{tab:main-results}, ``$f$ dist.'' is the $f$ distance and ``mAP'' the segmentation mAP (\%). The scalar $s$ is the RMS of the residual $r$ on the training set. The best values are in bold.}
  \label{tab:ablations}
  \footnotesize
  \setlength{\tabcolsep}{3pt}
  \begin{tabular}{lccccc}
    \toprule
    & \textbf{PSNR $\uparrow$} & \textbf{LPIPS $\downarrow$} & \textbf{FID $\downarrow$} & \textbf{{\boldmath$f$} dist.\ $\downarrow$} & \textbf{mAP $\uparrow$} \\
    \midrule
    \multicolumn{6}{l}{\textit{Source, with FIT}} \\
    $\mathcal{N}(0, I)$ & 19.80 & 0.197 & 5.14 & 0.060 & 61.4 \\
    $\mu_\phi(f)$ & \textbf{21.24} & 0.262 & 11.74 & 0.085 & 44.4 \\
    $\mathcal{N}(\mu_\phi(f), s^2 I)$ & 20.87 & 0.175 & 4.46 & 0.055 & 63.2 \\
    Diagonal $A$ & 20.81 & 0.176 & 4.50 & 0.055 & 63.4 \\
    \midrule
    \multicolumn{6}{l}{\textit{Velocity conditioning, with FUN}} \\
    None & 20.42 & 0.212 & 5.89 & 0.070 & 56.7 \\
    ControlNet & 20.73 & 0.179 & 4.55 & 0.056 & 63.9 \\
    \midrule
    \rowcolor{accent!12} \hspace{-\tabcolsep}{\color{accent}\vrule width 1.5pt}\hspace{2.5pt}\textbf{Ours} & 20.78 & \textbf{0.171} & \textbf{4.19} & \textbf{0.052} & \textbf{65.9} \\
    \bottomrule
  \end{tabular}
\end{table}

\Cref{tab:ablations} ablates the source and the velocity conditioning on ConvNeXt, changing one of the two at a time.
For the source, we compare FUN with the sources of \cref{sec:fit} ($\mathcal{N}(0, I)$, the mean alone and the mean with one noise level $s$) and with a FUN trained with only the diagonal of $A$.
PMRF~\citep{ohayon2025pmrf} suggests a small $s$; we use the RMS of the residual $r$ on the training set, the maximum-likelihood value of $s$.
The mean alone has the highest PSNR and the worst values on the other four metrics, consistent with the perception-distortion tradeoff~\citep{blau2018perception}.
The diagonal $A$ performs on par with the single noise level, and the sparse $A$ improves over both on every metric except PSNR, which suggests that the benefit of the predicted covariance lies in the correlations between latent values.
For the velocity, we compare FIT with a velocity that is not conditioned on $f$ and with a ControlNet~\citep{zhang2023controlnet} as in FeatInv.
Without conditioning, all five metrics degrade, in line with \cref{sec:fit}, and FIT outperforms ControlNet on all five.

\subsection{Toward Few-Step Inference}
\label{sec:few-step}

\begin{figure}[t]
\centering
\includegraphics{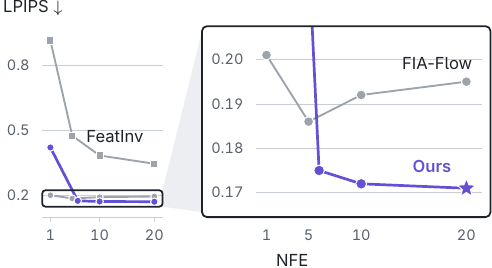}
\caption{ConvNeXt LPIPS at various NFEs for FeatInv, FIA-Flow and our method.
The best LPIPS is marked with a star.}
\label{fig:nfe-sweep}
\end{figure}

Our method remains a flow in the sense of \cref{sec:bg-flow}, since FIT is trained with the flow matching loss of \cref{eq:fit-loss} alone and \cref{alg:sampling} solves the resulting ODE numerically.
In principle, with the minimizer of \cref{eq:fit-loss}, the exact solution of this ODE carries the Gaussian of FUN onto $p_1(\cdot \mid f)$ (\cref{sec:fit}).
A solver with more steps comes closer to the exact solution, so we expect the samples to improve with the NFE.
The NFE is then a setting chosen at inference, and more compute can be spent on better samples without retraining.
\Cref{fig:nfe-sweep} supports this expectation: the LPIPS of our method decreases as the NFE grows (we use a single Euler step at NFE 1).
Its high value at NFE 1 is also expected, since for the minimizer in \cref{eq:fit-vstar} a single Euler step from $z_0$ returns the conditional mean of \cref{eq:mustar}.
FIA-Flow~\citep{ren2026fiaflow} is designed for a single step, its default setting, and at NFE 1 it is far ahead of our method.
In our shared setting, its LPIPS is lowest at a few steps and rises with more.
In practice fewer evaluations may suffice for our method, since the LPIPS at NFE 10 is very close to that at NFE 20 (a difference of 0.001).

In this work, we focus on the quality of the recovered images rather than on the speed of inference.
We argue that in practice feature inversion runs either occasionally, as when a viewer checks an alert (\cref{sec:intro}), or on the server, which in split inference has the compute that the device lacks.
Faster inference is nevertheless an advantage whenever it does not affect quality.
Because our method remains a flow, one route to fewer steps is reflow~\citep{liu2023rectified}.
We train a second velocity network, initialized again with SD2.1, using the loss of \cref{eq:fit-loss} on pairs $(z_0, \hat{z}_1)$, where $z_0$ is a draw from FUN and $\hat{z}_1$ is the solution that \cref{alg:sampling} computes from it.

\begin{table}[t]
  \centering
  \caption{Few-step results on ConvNeXt, where 2-RF denotes reflow. To allow odd NFEs, all rows use Euler's method. ``Hours'' is the total training time in A100 hours. The best values at each NFE are in bold.}
  \label{tab:reflow}
  \footnotesize
  \setlength{\tabcolsep}{2.5pt}
  \begin{tabular}{lcccccc}
    \toprule
    & \textbf{PSNR $\uparrow$} & \textbf{LPIPS $\downarrow$} & \textbf{FID $\downarrow$} & \textbf{{\boldmath$f$} dist.\ $\downarrow$} & \textbf{mAP $\uparrow$} & \textbf{Hours} \\
    \midrule
    \multicolumn{7}{l}{\textit{NFE 1}} \\
    FIA-Flow & 21.86 & 0.201 & 6.44 & \textbf{0.054} & \textbf{62.7} & 103 \\
    \rowcolor{accent!12} \hspace{-\tabcolsep}{\color{accent}\vrule width 1.5pt}\hspace{2.5pt}\textbf{Ours} & \textbf{21.87} & 0.421 & 50.75 & 0.105 & 38.2 & 56 \\
    \rowcolor{accent!12} \hspace{-\tabcolsep}{\color{accent}\vrule width 1.5pt}\hspace{2.5pt}\textbf{Ours + 2-RF} & 20.86 & \textbf{0.184} & \textbf{5.15} & 0.060 & 61.6 & 111 \\
    \midrule
    \multicolumn{7}{l}{\textit{NFE 5}} \\
    FIA-Flow & 21.20 & 0.186 & 4.90 & 0.055 & 62.3 & 103 \\
    \rowcolor{accent!12} \hspace{-\tabcolsep}{\color{accent}\vrule width 1.5pt}\hspace{2.5pt}\textbf{Ours} & \textbf{21.63} & 0.216 & 7.04 & \textbf{0.053} & \textbf{63.5} & 56 \\
    \rowcolor{accent!12} \hspace{-\tabcolsep}{\color{accent}\vrule width 1.5pt}\hspace{2.5pt}\textbf{Ours + 2-RF} & 20.73 & \textbf{0.179} & \textbf{4.61} & 0.057 & 63.3 & 111 \\
    \bottomrule
  \end{tabular}
\end{table}

\Cref{tab:reflow} compares the result (Ours + 2-RF) with FIA-Flow and with our method before reflow at the same NFEs.
With reflow, our method is competitive with FIA-Flow and outperforms it on the metrics of human perception.
The main drawback is that reflow increases the training time.
We expect other approaches to few-step flow matching~\citep{lee2024improving,geng2025mean,geng2026improved} to apply as well, which we leave to future work, and we expect our method to inherit future advances in flow matching in the same way.

\subsection{Defended and Compressed Features}
\label{sec:defended-compressed}

\begin{table}[t]
  \centering
  \caption{Feature inversion results on ConvNeXt with defended (DISCO) and compressed (JPEG) features. The original top-1 accuracy of ConvNeXt is 86.4\%, and each head gives the change. The best values are in bold.}
  \label{tab:distortions}
  \footnotesize
  \setlength{\tabcolsep}{3pt}
  \resizebox{\linewidth}{!}{%
  \begin{tabular}{lr>{\hspace{3pt}}ccccc>{\hspace{3pt}}ccccc}
    \toprule
    & & \multicolumn{5}{c}{\textbf{DISCO} ($\lambda = 0.4$, top-1 $-2.0$\%)} & \multicolumn{5}{c}{\textbf{JPEG} ($Q = 3$, top-1 $-0.8$\%)} \\
    \cmidrule(lr){3-7} \cmidrule(lr){8-12}
    \textbf{Method} & \textbf{NFE} & \textbf{PSNR $\uparrow$} & \textbf{LPIPS $\downarrow$} & \textbf{FID $\downarrow$} & \textbf{{\boldmath$f$} dist.\ $\downarrow$} & \textbf{mAP $\uparrow$} & \textbf{PSNR $\uparrow$} & \textbf{LPIPS $\downarrow$} & \textbf{FID $\downarrow$} & \textbf{{\boldmath$f$} dist.\ $\downarrow$} & \textbf{mAP $\uparrow$} \\
    \midrule
    DO & 1 & 17.99 & 0.735 & 146.37 & 0.223 & \phantom{0}4.8 & 17.75 & 0.736 & 156.85 & 0.202 & \phantom{0}3.3 \\
    DB & 1 & 17.38 & 0.769 & 175.73 & 0.332 & \phantom{0}2.7 & 17.82 & 0.736 & 159.34 & 0.221 & \phantom{0}4.4 \\
    FIA-Align & 1 & 19.39 & 0.509 & \phantom{0}78.26 & 0.121 & 22.0 & \textbf{18.11} & 0.645 & 128.08 & 0.170 & \phantom{0}7.4 \\
    FIA-Flow & 1 & \textbf{19.51} & 0.260 & \phantom{00}8.86 & 0.057 & 55.0 & 17.97 & 0.364 & \phantom{0}22.33 & 0.096 & 34.2 \\
     & 5 & 19.19 & 0.254 & \phantom{00}7.25 & 0.059 & 53.4 & 17.91 & 0.387 & \phantom{0}19.62 & 0.097 & 34.0 \\
     & 20 & 18.92 & 0.266 & \phantom{00}8.20 & 0.067 & 47.4 & 17.69 & 0.407 & \phantom{0}23.03 & 0.109 & 28.4 \\
    FeatInv & 20 & 13.15 & 0.462 & \phantom{0}13.35 & 0.074 & 44.0 & 12.82 & 0.536 & \phantom{0}19.56 & 0.108 & 27.5 \\
    \rowcolor{accent!12} \hspace{-\tabcolsep}{\color{accent}\vrule width 1.5pt}\hspace{2.5pt}\textbf{Ours} & 20 & 18.68 & \textbf{0.227} & \phantom{00}\textbf{5.61} & \textbf{0.053} & \textbf{60.2} & 16.78 & \textbf{0.353} & \phantom{0}\textbf{10.28} & \textbf{0.090} & \textbf{37.7} \\
    \bottomrule
  \end{tabular}}
\end{table}

In practical split inference, the features are often altered before they are sent, so we also study our method on such features.
For example, a defense may be applied to protect the image~\citep{mireshghallah2020shredder,azizian2024privacy,xia2025theoretical}, or the features may be lossily compressed for transmission~\citep{choi2018deep,eshratifar2019bottlenet}.
Both are expected to reduce the information that the features carry about the image.
\Cref{tab:distortions} compares the methods on ConvNeXt for features defended with DISCO~\citep[prune ratio $\lambda = 0.4$]{singh2021disco} or compressed with JPEG~\citep[quality $Q = 3$]{wallace1992jpeg}, where the top-1 accuracy of the network degrades by no more than 2.0\%.
All methods are trained and tested on these features.
In both scenarios our method keeps the leads observed in \cref{tab:main-results}, in most cases increasing them compared to the original features.
The growth is largest in FID, where the lead over the best baseline rises from 0.71 on the original features to 1.64 with DISCO and 9.28 with JPEG.
We attribute this to how strongly our method is conditioned on $f$, which matters more when the features carry less information and more of the image has to be generated.
\Cref{fig:main-examples} (middle and bottom rows) compares the images recovered from defended and from compressed features.

\section{Conclusion}
\label{sec:conclusion}

We presented a method for black-box feature inversion in which FUN predicts a Gaussian from the features and FIT carries a draw from it to an image by flow matching conditioned on the same features.
On five networks, the recovered images score best on most metrics of human perception and on utility for vision models (\cref{tab:main-results}), with these leads retained and increased on defended and compressed features (\cref{tab:distortions}).
Both stages are trained in the latent space without decoding or a perceptual loss, and the LPIPS of the recovered images is still the lowest on every network.
The vision models behind the utility metrics have never seen the network or its features, and on the recovered images they largely reproduce their own classification, segmentation and vision-language outputs on the originals (\cref{tab:main-results}).
A server that receives only the features can therefore add off-the-shelf vision models for new tasks without a change to the network on the device.
We believe that our work demonstrates and advances the practical use of feature inversion for humans and vision models.

\paragraph{Limitations.}
Our method assumes features that share the layout of the image (\cref{sec:bg-inversion}).
This is the common case in split inference with vision models~\citep{choi2018deep,choi2018near,eshratifar2019bottlenet,dong2022privacy,lei2025drag,zhang2025unlocking,ren2026fiaflow} and holds for the convolutional networks and vision transformers we test, but not for arbitrary features.
Sampling also takes several function evaluations, although the NFE is chosen at inference.
Since our method remains a flow, reflow is one route to fewer evaluations (\cref{sec:few-step}), and we believe that future advances in flow matching will reduce the NFE further.
We run our experiments on ImageNet at $512^2$ and leave higher resolutions and datasets from specialized domains to future work.

\paragraph{Ethics.}
While we believe that feature inversion has beneficial uses, we recognize that our method may be used as an attack, as discussed in previous works~\citep{he2019model,dong2022privacy,yang2022measuring,yin2023ginver,li2023gan,chen2024dia}.
Especially given the results on defended features (\cref{sec:defended-compressed}), we encourage the use of this work to inform better practices in split inference leading to improved and fortified defenses against state-of-the-art inversion methods.

\subsection*{AI Use Statement}
We used Claude (Anthropic) in the writing and in parts of the research.
It helped us draft and edit text, search for related work, write code and make figures.
We reviewed all AI-assisted output and take full responsibility for this work.

\bibliographystyle{iclr2027_conference}
\bibliography{main}

\end{document}